\documentclass{book}    
\usepackage{piers}  
\usepackage{booktabs}
\begin{document}

\title{Improved Active Fire Detection using Operational U-Nets}
\maketitle

\author      {F. M. Lastname}
\affiliation {University}
\address     {}
\city        {Boston}
\postalcode  {}
\country     {USA}
\phone       {345566}    
\fax         {233445}    
\email       {email@email.com}  
\misc        { }  
\nomakeauthor

\author      {F. M. Lastname}
\affiliation {University}
\address     {}
\city        {Boston}
\postalcode  {}
\country     {USA}
\phone       {345566}    
\fax         {233445}    
\email       {email@email.com}  
\misc        { }  
\nomakeauthor

\begin{authors}

{\bf Ozer Can Devecioglu}$^{1}$, {\bf Mete Ahishali}$^{1}$, {\bf Fahad Sohrab}$^{1}$ {\bf Turker Ince}$^{2}$ {\bf and Moncef Gabbouj}$^{1}$\\
\medskip
$^{1}$Department of Computing Sciences, Tampere University, Tampere, Finland\\
$^{2}$Department of Electrical and Electronics Engineering, Izmir University of Economics, Izmir, Turkey
\end{authors}

\begin{paper}

\begin{piersabstract}
As a consequence of global warming and climate change, the risk and extent of wildfires have been increasing in many areas worldwide. Warmer temperatures and drier conditions can cause quickly spreading fires and make them harder to control; therefore, early detection and accurate locating of active fires are crucial in environmental monitoring. Using satellite imagery to monitor and detect active fires has been critical for managing forests and public land. Many traditional statistical-based methods and more recent deep-learning techniques have been proposed for active fire detection. In this study, we propose a novel approach called Operational U-Nets for the improved early detection of active fires. The proposed approach utilizes Self-Organized Operational Neural Network (Self-ONN) layers in a compact U-Net architecture. The preliminary experimental results demonstrate that Operational U-Nets not only achieve superior detection performance but can also significantly reduce computational complexity. 
\end{piersabstract}

\psection{Introduction}
Wildfires may occur due to many reasons, including fires started by lightning, unauthorized human activity, and global warming. Forest fires negatively affect ecosystems over the long run by disrupting vegetation dynamics, emitting greenhouse gases, changing the habits of wildlife, and removing the land cover. Because of climate change and land-use change, wildfires are expected to become increasingly frequent and intense, with a global rise in extreme fires of up to 14\% by 2030, 30\% by the end of 2050, and 50\% by the end of the century\footnote{https://www.unep.org/ https://www.grida.no/}. Therefore, early detection and management of wildfires are critical. Determining the precise location of an active fire is one of the most effective ways of managing and monitoring, which makes it crucial for firefighting activities. Modern Remote Sensing (RS) technologies have been commonly used to monitor and examine the earth's surface and land cover on a wide range of scales in recent years. Many applications of forest monitoring, including the mapping of burned areas, are made possible by the availability of high spatiotemporal resolution data and multispectral imagery from the RS systems. 

Avgerinakis et al. \cite{avgerinakis2012smoke}, evaluated the smoke candidate blocks by using the histograms of oriented gradients (HOGs) and histograms of optical flow (HOFs), accounting for both appearance and motion data. A support vector machine (SVM) classifier is employed to determine fire regions. Barmpoutis et al. \cite{barmpoutis2013real} introduced a new feature with the intention of estimating the spatiotemporal consistency energy of videos. Their proposed feature classifier improved the robustness of the fire detection system. Gunay et al., \cite{gunay2010fire} used hidden Markov models to separate flame flicker from the motion of moving objects that were fire-colored in movies (HMMs). To detect active fire in Landsat-8 datasets at the pixel level, Rostami et al. \cite{rostami2022active} introduced a deep convolutional neural network (CNN) "MultiScale-Net". They used dilated convolutions and convolutional kernels with multiple sizes. SWIR2, SWIR1, and Blue bands were also used as input to achieve the F1-score and IoU of 91.62\% and 84.54\%, respectively. Kang et al. \cite{kang2022deep}, proposed an ensemble learning method using both Random Forest Classifier and CNNs. They obtained overall accuracy, precision, recall, and F1-score as 0.98, 0.91, 0.63, and 0.74, respectively. Cheng et al. \cite{cheng2019smoke}, proposed a smoke detection system using both Deeplabv3+ and Generative Adversarial Network (GAN). They used Deeplabv3+ to identify the smoke pixels, and GAN was utilized to produce segmentation maps of these regions.

Most of the above-mentioned methods in the literature either used hand-crafted features with relatively traditional learning models or directly processed raw image data using computationally complex deep learning models. In practice, the usage of such deep networks is unfeasible in near real-time with a low-power computational environment. The latter approach with deep network models is known to suffer from low generalization performance. To address these limitations, we employed novel Self-Operational Neural Networks (Self-ONNs) \cite{kiranyaz2020operational,kiranyaz2021exploiting,malik2020fastonn,kiranyaz2021self} as a heterogeneous network model containing non-linear neurons. In our earlier works, Self-ONNs \cite{devecioglu2021real,ince2021early, kiranyaz2022blind,malik2021real}, have been utilized to achieve superior performance for challenging real-world problems. As a consequence, in this study, we propose compactly optimized Operational U-Nets containing Self-ONNs in internal layers of both encoder and decoder architectures for active fire detection in satellite images and evaluate their performance extensively over the benchmark Landsat-8 \cite{de2021active} dataset.

\psection{Methods}
\psubsection{Self-Operational Neural Networks}
In this section, we briefly review the Self-ONNs and some of their key characteristics. Different from the convolution operator of CNNs, the nodal operator of each generative neuron of a Self-ONN
can perform any nonlinear transformation which can be expressed based on Taylor approximation around the near origin:

\begin{equation}
\psi(x)= \sum_{n=0}^\infty \frac{\psi^n 0}{n!} x^n
\label{eq1}
\end{equation}
which can approximate any arbitrary function $\psi$ well near 0.
When the activation function bounds the neuron’s input feature maps in the vicinity of 0, e.g., hyperbolic tangent (tanh) function, the formulation of (1) can be exploited to form a composite nodal operator. If we denote $\frac{\psi^n0}{n!}$ as $w_n$:

\begin{equation}
\psi(w,y)=w_0+w_1y^1+w_2y^2+...+w_Qy^Q
\label{eq2}
\end{equation}
where $w_0$ is the bias, and $w_1-w_Q$ are the Mclauren coefficients, which will be computed through the backpropagation. Further information regarding Self-ONNs' theory and forward propagation formulations can be found in \cite{kiranyaz2020operational}.

\psubsection{Active Fire Detection using Operational U-Nets}
The general framework of our proposed active fire detection scheme is shown in Figure \ref{im1}. Similar to \cite{de2021active}, we form an RGB image by selecting the same three channels (c7, c6, and c2) from the multispectral data consisting of 10 frequency bands. Then, each channel of the input image X is resized to 256×256 and linearly scaled to a range of [-1,1] as follows,

\begin{equation}
X_N\left(i,j\right)=2\frac{X\left(i,j\right)-X_{min}}{X_{max}-X_{min}}-1
\label{eq3}
\end{equation}

The overall framework of the proposed method is given in Figure \ref{im1}. The proposed Operational U-Net network consists of three main parts: encoder, decoder, and skip connections. The encoder network aims to learn useful features. The decoder network is connected to the encoder with skip connections, and it generates the output semantic segmentation mask using learned features. Skip connections are shortcut connections helping the indirect flow of gradients from the early layers of the encoder to the decoder. In this way, we aim to improve the overall quality of the produced segmentation masks. In the proposed approach, convolutional layers are replaced with operational layers in order to further improve the mapping capability of the proposed framework. 

\begin{figure}[hb]
    \centering
    \includegraphics[width=1\linewidth]{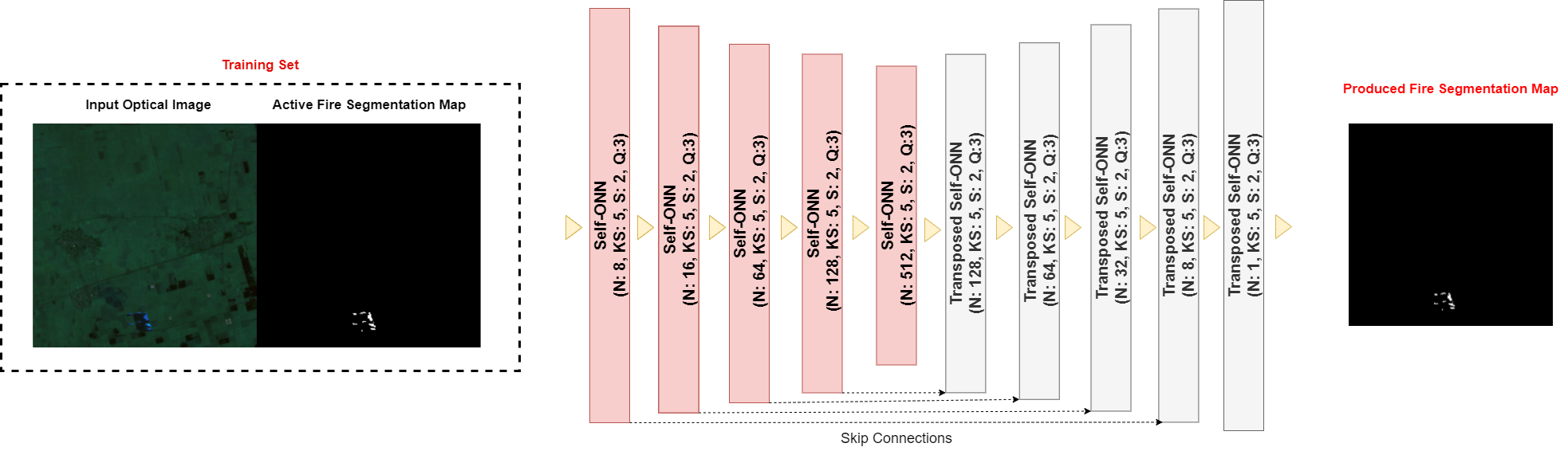}
    \caption{The general framework of improved active fire detection using Operational U-Nets}
    \label{im1}
\end{figure}

The Operational U-Net architecture contains 5 operational layers as well as 5 transposed operational layers with skip connections, as shown in Figure \ref{im1}. The kernel sizes are set to 5, with the exception of the last transposed layer, which has a kernel size of 6. For both operational and transposed operational layers, the stride is set to 2. The nonlinear activation function tanh is utilized. For all layers, Q is set to 3. We use a training scheme with a batch size of 8 and a maximum of 1000 epochs for all runs. Adam's optimizer\cite{kingma2014adam} is utilized in the training with a learning rate of 1e-5.

\psection{Landsat-8 Dataset}
The Landsat-8 satellite, launched in 2013 by NASA, is equipped with the Operational Land Imager (OLI) and Thermal Infrared Sensor (TIRS) instruments that capture high-resolution multispectral imagery of the Earth's surface. Land cover, vegetation health, water resources, and natural disasters are just a few of the topics covered in depth by the Landsat-8 collection. A long-term record of changes in the Earth's surface through time is provided by the satellite's orbit, which enables global coverage every 16 days. 

For training and testing of our proposed model, the processed version of the benchmark Landsat-8 dataset is used for all experiments \cite{de2021active}. As preprocessing, Landsat-8 images are first split to image patches with 256x256 pixels. After normalization, the dataset is split into a train set, validation set, and test set with proportions of 40\%, 10\%, and 50\%, respectively.
\psection{Results}
The results of the proposed method and benchmark models in terms of precision, recall, IoU, and F1-score are reported in Table \ref{results}. Figure \ref{im2} displays the corresponding input images, along with the segmented outputs produced by the Operational U-Net model. 

To evaluate the effectiveness of the proposed model, we compared it with several benchmark models. These included compact U-Net and deep U-Net models with both 10-channel and 3-channel input data. To further improve the performance of the deep U-Net models, we also applied transfer learning by replacing their encoder parts with pre-trained DenseNet121, ResNet50, Inception-v3, and MobileNet models on mageNet\cite{deng2009imagenet}. Fine-tuning was then applied using the same training procedure as that used for the proposed approach. The results of our experiments are shown in Table \ref{results}, where we report the F1 score along with other performance metrics. We found that the Operational U-Net outperformed all the benchmark models with an F1 score of 90.2\%. This result indicates the superior performance of our proposed model in the task of image segmentation.

When considering the scarcity of positive situations in the dataset, the proposed method increased the generalization and decreased the false positive rate against the deep network, achieving the highest f1 score with higher precision. The proposed model outperforms deep transfer learning models by 12-14\% in terms of F1-Score. Our compact network decreases the false positive rate. The Operational U-Net achieves a performance gap of approximately 3\% in the F1 score over both deep and light Convolutional U-Net models from \cite{de2021active}. Additionally, it can achieve even 1\% better than deep Convolutional U-Net when they use 10 channels of input data. Note that the proposed approach is computationally efficient with the smallest number of trainable parameters compared to other approaches.

\begin{table}[]
\label{results}
\begin{tabular}{@{}ccccccc@{}}
\toprule
\textbf{Method}               & \textbf{Channels} & \textbf{Precision (\%)} & \textbf{Recall (\%)} & \textbf{IoU (\%)}  & \textbf{F1-score (\%)} & \textbf{Parameters(M)} \\ \midrule
U-Net {[}9{]}       & 10       & 84.6      & 94.1   & 80.3 & 89.1     & 34.5       \\
U-Net {[}9{]}       & 3        & 84.2      & 90.6   & 77.4 & 87.3     & 34.5       \\
U-Net-Light {[}9{]} & 3        & 76.8      & 93.2   & 72.7 & 84.2     & 2.2        \\
DenseNet121          & 3        & 86.6      & 68.1   & 61.6 & 76.2     & 12.0       \\
ResNet50             & 3        & 84.2      & 69.2   & 61.3 & 76.0     & 32.5       \\
Inception-v3         & 3        & 84.2      & 68.7   & 60.9 & 75.7     & 29.8       \\
MobileNet-v2         & 3        & 82.0      & 73.6   & 63.4 & 77.6     & 8.0        \\
\textbf{Op-UNet}     & 3        & 98.7      & 83.1   & 82.1 & 90.2     & 4.3 \\ \bottomrule
\end{tabular}
\caption{Performance of the proposed \textbf{(Op-UNet)} method and benchmark models}
\end{table}

\begin{figure}[h!]
    \centering
    \includegraphics[width=0.95\linewidth]{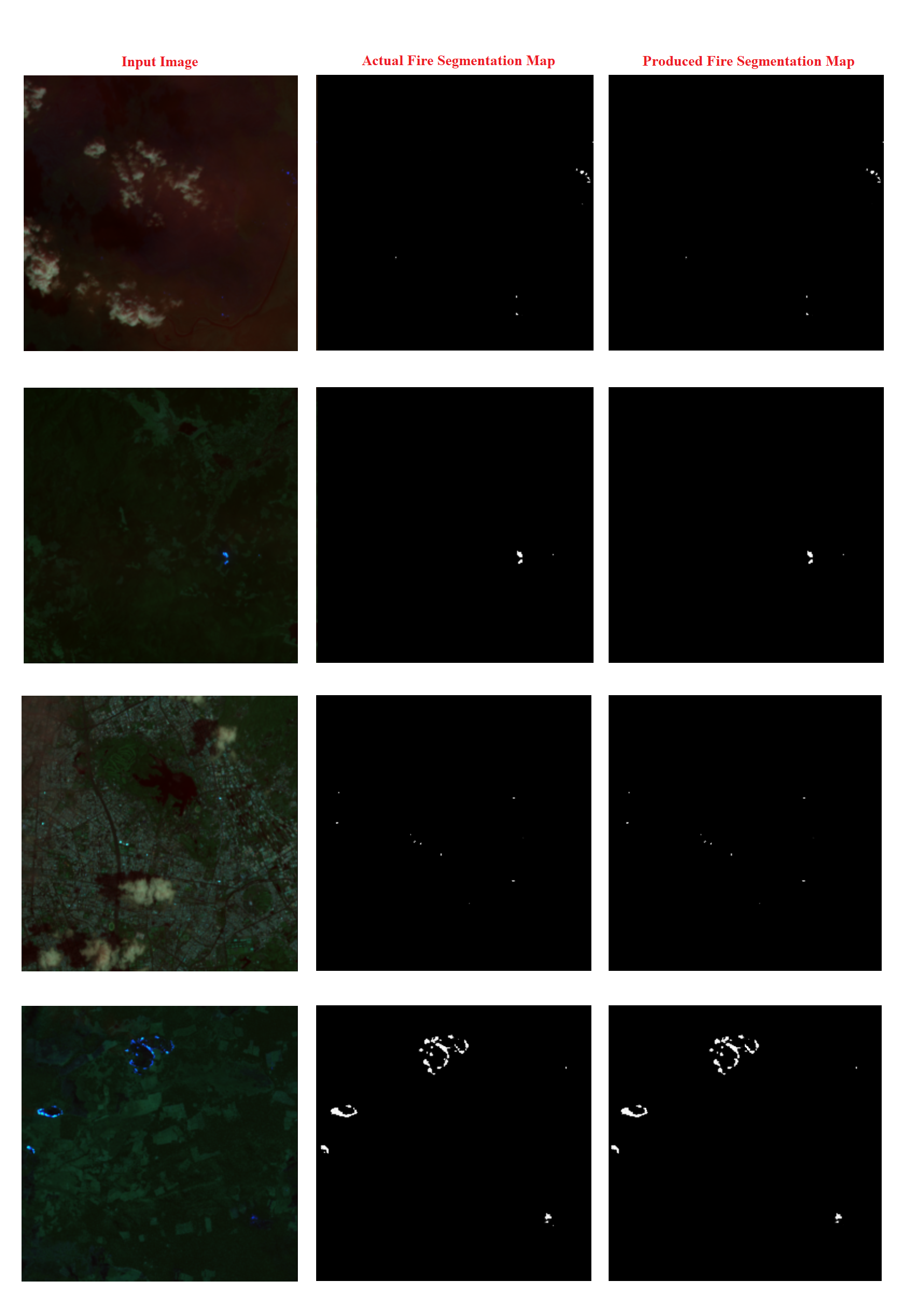}
    \caption{Corresponding Operational U-Net segmented maps and their input Landsat-8 images.}
    \label{im2}
\end{figure}

\psection{Conclusion}
In this study, we propose a new method based on the Operational U-Nets to detect active fire regions in Landsat-8 imagery. Experimental results using a collection of large labeled images show that the proposed approach achieves superior accuracy compared to the deeper models. Moreover, the proposed model has significantly less computational complexity than the competing methods, making it feasible for near real-time implementations (such as on-board processing of satellite data).

As a potential alternative to our current approach, we aim to explore one-class classification techniques \cite{sohrab2023graph,sohrab2020ellipsoidal} in the future for active fire detection. These approaches can effectively handle the data scarcity issue in detecting positive class samples, and they have the potential to enhance the performance of our proposed method further. By investigating these techniques, we hope to make significant contributions to the field of active fire detection.
\ack
This work was supported by the NSF-Business Finland project AMALIA. Foundation for Economic Education (Grant number: 220363) funded the work of Fahad Sohrab at Haltian.

\bibliography{refs}
\bibliographystyle{unsrt}

\end{paper}

\end{document}